%% file: main.tex
\documentclass[10pt,twocolumn,letterpaper]{article}

\usepackage[pagenumbers]{cvpr} 

\input{preamble}

\definecolor{cvprblue}{rgb}{0.21,0.49,0.74}
\usepackage[pagebackref,breaklinks,colorlinks,citecolor=cvprblue]{hyperref}
\usepackage{multirow}
\usepackage{makecell}
\usepackage{xcolor,colortbl}
\usepackage{color}
\usepackage{comment}
\usepackage[misc]{ifsym}

\definecolor{minetable1colorx}{rgb}{0.75, 0.75, 0.75}

\definecolor{lightgrey}{gray}{0.9}

\usepackage{algorithm}
\usepackage{algorithmic}

\title{VLM-Assisted Continual learning for Visual Question Answering in Self-Driving}

\author{%
\textbf{Yuxin Lin, Mengshi Qi, Liang Liu, Huadong Ma} \\ 
State Key Laboratory of Networking and Switching Technology~~~\\ Beijing University of Posts and Telecommunications, China\\
\{linyuxin2019, qms, liangliu, mhd\}@bupt.edu.cn
}

\begin{document}
\maketitle

\begin{abstract}
In this paper, we propose a novel approach for solving the Visual Question Answering (VQA) task in autonomous driving by integrating Vision-Language Models (VLMs) with continual learning. In autonomous driving, VQA plays a vital role in enabling the system to understand and reason about its surroundings. However, traditional models often struggle with catastrophic forgetting when sequentially exposed to new driving tasks, such as perception, prediction, and planning, each requiring different forms of knowledge. To address this challenge, we present a novel continual learning framework that combines VLMs with selective memory replay and knowledge distillation, reinforced by task-specific projection layer regularization. The knowledge distillation allows a previously trained model to act as a "teacher" to guide the model through subsequent tasks, minimizing forgetting. Meanwhile, task-specific projection layers calculate the loss based on the divergence of feature representations, ensuring continuity in learning and reducing the shift between tasks. Evaluated on the DriveLM dataset, our framework shows substantial performance improvements, with gains ranging from $20.11\%$ to $35.16\%$ across various metrics. These results highlight the effectiveness of combining continual learning with VLMs in enhancing the resilience and reliability of VQA systems in autonomous driving. We will release our source code.
\end{abstract}
%
\section{Introduction}

With the rapid advancement of technology, autonomous driving~\cite{aitech,autonomoustechnology}  has become a landmark achievement in the development of modern transportation systems. Self-driving cars utilize various advanced sensors, cameras, and artificial intelligence algorithms to interpret the traffic environment, promising a safer and more efficient driving experience. 
Nowadays, Vision-language models~\cite{EM-VLM4AD,Drivevlm,clip,blip,llava}, play a crucial role as the core of autonomous driving systems. These models understand and predict road conditions and their complexity by parsing visual images and video data collected from camera and other sensors, combined with simultaneously processed verbal commands. 

\begin{figure*}[!t]
    \centering
    \includegraphics[width=0.9\linewidth]{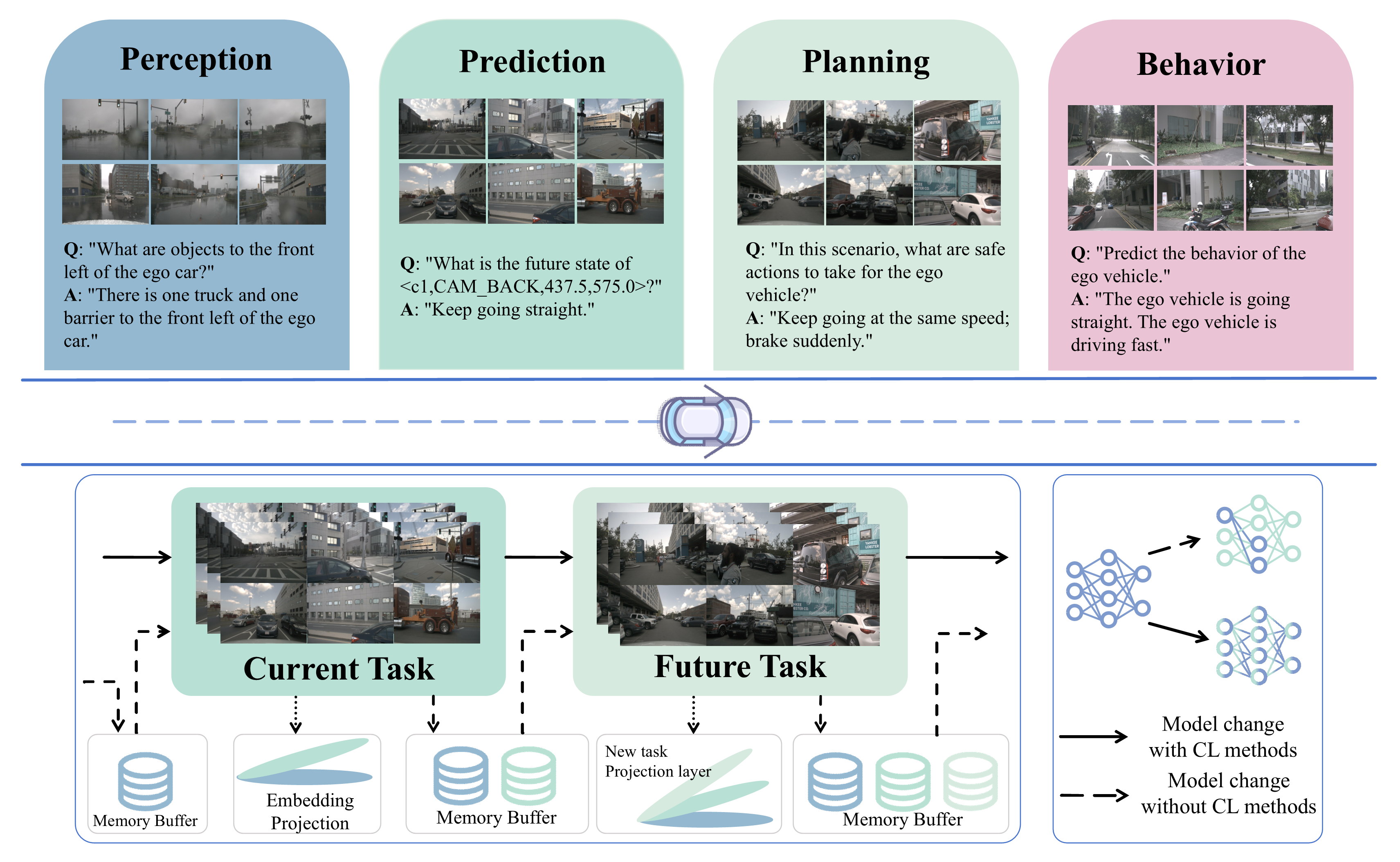}
    \caption{This illustration presents a framework for a Vision-Language Model (VLM) designed for multi-task autonomous driving. (a) The upper section shows four essential tasks: Perception, Prediction, Planning, and Behavior, each represented with a specific question-answer example to demonstrate the model's response to various driving scenarios. (b) The lower section outlines the model's simplified pipeline, incorporating memory replay with knowledge distillation and projection layer regularization to enhance continual learning capabilities across tasks. (c) The diagram in the bottom-right corner illustrates how continual learning methods enable the model to add new knowledge and functionality without overwriting or erasing prior knowledge.}
    \label{structure}
\end{figure*}

Although vision-language models (VLMs)~\cite{EM-VLM4AD,Drivevlm} show great potential in handling complex driving tasks, these models still face multiple challenges in autonomous driving applications. With the diversity of task types including perception, prediction, and planning, VLMs in autonomous driving increasingly need to be equipped with strong continual learning capabilities. Continual learning~\cite{DER,EWC,EWC++,MER,vqacl} allows models to accumulate new knowledge based on existing knowledge without having to go back to initial training data to relearn. Currently, continual learning is also introduced in a number of VLM models and VQA models in mitigating catastrophic forgetting~\cite{vqacl,lei2023symbolic} in different domains, e.g., medical domain~\cite{yu2024boosting,bai2023revisiting,chen2024llm,yi2023towards} and so on. 
However, in the context of autonomous driving, the application of these methods requires special adaptation and optimization. Any learning inaccuracies may lead to serious safety consequences. Addressing catastrophic forgetting within the VLM framework for autonomous driving tasks—such as perception, prediction, and planning—requires innovative strategies beyond standard sequential training. As illustrated in Figure~\ref{structure}, VLMs trained on new tasks may experience catastrophic forgetting, where prior task knowledge is overwritten, leading to loss of critical information. Therefore, ensuring that VLMs are able to adapt to new driving environments and tasks without losing prior knowledge is crucial for improving the reliability and safety of autonomous driving systems.

To address above-mentioned issues, in this work, we introduce a novel hybrid continual learning strategy with VLMs, which combines selective memory replay~\cite{ER} with knowledge distillation~\cite{KD}, enhanced by projection layer regularization. Specially, selective memory replay periodically reinforces previously learned information to mitigate forgetting, while knowledge distillation enables the model trained on earlier tasks to act as a teacher, guiding subsequent tasks and preserving complex task knowledge. This strategy balances learning new tasks and retaining essential prior knowledge, crucial in a continually changing autonomous driving environment. By replaying real driving data, the system retains critical safety-related information across diverse driving conditions, thus enhancing the model's reliability and generalization capability.

To further strengthen our model against the challenge of catastrophic forgetting, we introduce a new task-specific regularization methods of projection layers. In models with a relatively large number of model parameters and experimental data, both the classical continual learning parameter regularization approach~\cite{EWC} and the gradient projection approach~\cite{GEM} face challenges in terms of computational resource requirements and operational efficiency. Therefore, following the idea of regularization, we regularize the learning process by carefully designing the model feature projection layer and calculating the loss based on the differences in feature representations in successive tasks. This regularization not only reduces additional computational and memory requirements, but also prevents the dilution of previously learned features and ensures that the model's learning trajectory closely follows the continuum, thus mitigating the effects of sudden knowledge transfer.

Our main contributions can be summarized as follows:  
\begin{enumerate}
    \item We introduce a new framework for tackling the Visual Question Answering (VQA) task in autonomous driving by leveraging Vision-Language Models (VLMs) in combination with continual learning strategies, enabling the model to adapt seamlessly across different driving tasks.
    \item We present an innovative approach combining memory replay and knowledge distillation, and enhance the continual learning by introducing a projection layer to achieve regularization in the feature embedding layer.
    \item We benchmark our model on the DriveLM dataset, achieving significant performance improvements. Compared to the baseline model, our method demonstrated improvements across various metrics, ranging from a minimum of $20.11\%$ to a maximum of $35.16\%$, highlighting its superior capacity to handle complex multimodal VQA tasks in autonomous driving environments.
\end{enumerate}

\section{Related Work}

\noindent\textbf{Vision Language Models for VQA.} In the field of autonomous driving, the application of Vision-Language Models (VLMs) is rapidly expanding~\cite{you2024v2x,Drivevlm}, enhancing the system's understanding of and decision-making capabilities in complex driving environments. VLMs, through extensive image-text pre-training, have already provided the ability for zero-shot learning in autonomous driving~\cite{vlmsurvey}. For example, the DriveVLM~\cite{Drivevlm} system explores the integration of VLMs within traditional autonomous driving technologies, enhancing the vehicle's spatial reasoning and planning capabilities. Additionally, the DriVLMe~\cite{drivlme} and Co-driver~\cite{Co-driver} projects explore how LLMs mimic human understanding and behavior in handling complex road conditions from the perspectives of experiential learning and social interaction, respectively. In recent years, Vision-Question-Answering (VQA) datasets for autonomous driving, such as the Nuscenes-QA~\cite{nuscenesqa}dataset, DriveLM dataset~\cite{drivelm}, have further facilitated advancements in the multimodal understanding of driving scenarios. Meanwhile, topology-aware reasoning has also emerged as an important direction. For instance, T2SG~\cite{lv2025t2sg} introduces a traffic-topology scene graph to explicitly model relational structures in driving environments, providing complementary insights for question answering systems. Other efforts enhance driving VLMs through auxiliary predictive tasks, such as uncertainty-guided attention prediction~\cite{zhu2023unsupervised} or knowledge-graph-based retrieval augmentation~\cite{ye2025safedriverag}, both of which improve robustness and interpretability in complex driving scenes. In our model, we delve deeper into the field of continual learning for autonomous driving, building upon the EM-VLM4AD~\cite{EM-VLM4AD} framework.  

\noindent\textbf{Continual Learning.} Continual learning (CL) techniques like memory replay~\cite{DER}, regularization~\cite{EWC,MAS}, and optimization~\cite{GEM,OGD} mitigate catastrophic forgetting by balancing new and old tasks. Memory replay stores key data or features in a buffer, using methods such as random selection~\cite{chaudhry2019tiny} or feature averaging~\cite{iCaRL}. Regularization approaches add terms that selectively preserve crucial parameters based on their importance, assessed using tools like the Fisher Information Matrix (FIM)~\cite{EWC,EWC++}. Optimization methods, such as OGD~\cite{OGD}, maintain previous gradient directions while adjusting current gradients for orthogonality. However, in our approach, we employ some techniques for text vectorization and clustering of different tasks in the dataset to optimize data selection in the memory replay mechanism. In addition to this, we naturally combine memory replay with knowledge distillation (KD), which additionally integrates the past information of the old model. Due to the relatively large number of parameters in the VLM, regularization with gradient projection via parametric aspects is not very practical. Therefore, we emulated the idea of regularization and projection. We did this by comparing the projection layer gaps between different tasks after the model's feature embedding was introduced. Additionally, we added extra terms to the loss function to further enhance the continual learning effect at the model's feature level.

\section{Proposed Approach}\label{sec3}
\begin{figure*}[htbp]
  \centering
  \begin{subfigure}{.473\textwidth}
    \centering
    \includegraphics[width=\linewidth]{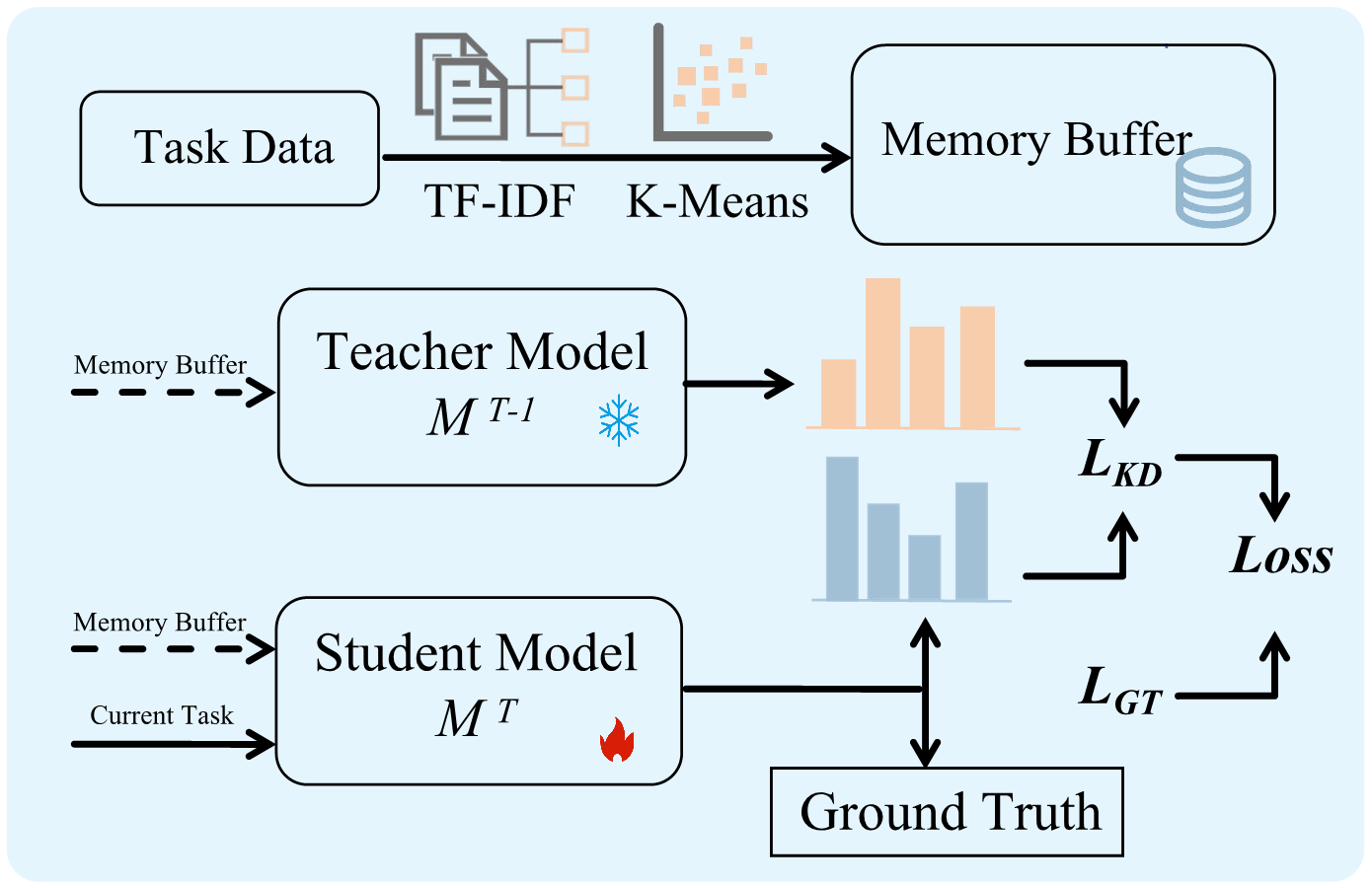}  
    \caption{Memory replay with knowledge distillation}
    \label{ERKD}
  \end{subfigure}\hfill
  \begin{subfigure}{.527\textwidth}
    \centering
    \includegraphics[width=\linewidth]{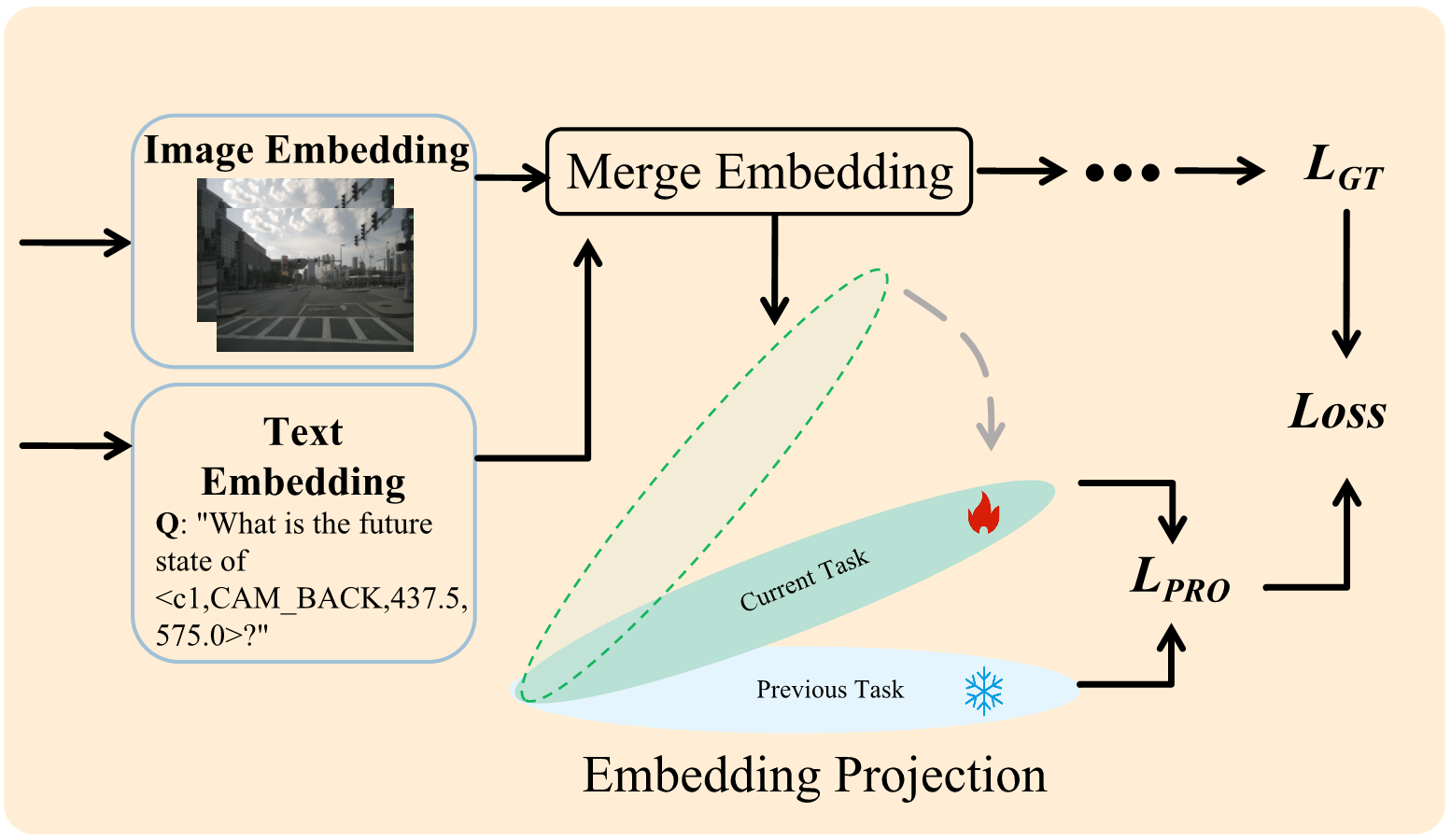}  
    \caption{Task-specific Embedding projection regularization}
    \label{Pro}
  \end{subfigure}\hfill
  \caption{(a)~\textbf{Memory replay with knowledge distillation} shows memory replay and knowledge distillation when Memory data coming in. Optimizing data selection using TF-IDF and K-means clustering to maintain a diverse and representative memory set. (b)~\textbf{Task-specific embedding projection regularization} demonstrates task-specific projection layers within an autonomous driving model to maintain feature continuity and mitigate catastrophic forgetting by transforming multimodal embedding into unique task-specific spaces and regulating the model with a specialized loss function.}
\end{figure*}

\subsection{Overview}
Targeting the autonomous driving visual question answering (VQA) dataset, we propose a new continual learning approach for Vision-Language Model (VLMs). This method alternates continual learning in two phases: one part is carried out during the data replay phase, and the other part during the current task data training phase. We construct a simple autonomous driving VLM using a straightforward image embedding network (ViT)~\cite{ViT} combined with a pre-trained T5 language model, applying the designed continual learning method to this VLM. The visualization of our continual learning model is shown in Figure~\ref{structure}, which includes memory replay with knowledge distillation and embedding projection. This unified design offers a coordinated solution to the stability–plasticity trade-off in continual multimodal learning, achieving complementary effects unattainable with isolated strategies. Next, we will delve into the details of each continual learning phase.

\subsection{Memory replay with Knowledge Distillation}
In multitask data training for autonomous driving, models often suffer from catastrophic forgetting when confronted with different tasks. To address this challenge, we introduce an new memory replay with knowledge distillation approach(Figure~\ref{ERKD}), to retain important historical data to help the model learn new tasks while efficiently maintaining the memory of old tasks and avoiding knowledge loss.    
 
Specially, our approach divides the data into $D = \{D_1, D_2, \ldots, D_n\}$ based on tasks \(T_n\), where each task is a different type of VQA dataset, and each task has a corresponding multi-class problem. Different from randomly selecting the past task data to compose the memory, in our approach, we choose the combination of TF-IDF and K-means to record the memory as the following.

\noindent\textbf{Memory replay.} Firstly, we perform text vectorization on the data in the task and use TF-IDF to extract features from the problem set, which helps us understand the main semantic content of the data. 
Then, by applying the K-means algorithm, we cluster the questions into several categories, which ensures that the data stored in the memory is representative and diverse in terms of contents.
For each clustering, we refine the selection process: the number of samples to be selected for each category, denoted by \(S_{c,n}\), is determined based on the proportion of data in each category, ensuring that the importance of each category is balanced. Specifically, we calculate the ratio of the total data volume \(D_n\) to the data volume \(D_{c,n}\) in each category \(c\) within task \(D_n\) , and use this ratio to determine \(S_{c,n}\). This approach not only improves the coverage of the samples but also increases the adaptability and robustness of the model to different types of problems. \(S_{c,n}\) can be formulated as the following:
\begin{equation}
S_{c,n} = \left\lfloor \frac{|D_{c,n}|}{\sum_{i=1}^{k_n} |D_{i,n}|} \times S_n \right\rfloor,
\end{equation}
where \( D_{c,n} \) represents the volume of data for category \( c \) in task \( n \), \( k_n \) represents the total number of categories in that task, and \( S_n \) represents the total number of samples chosen.

Eventually, a specified number of data samples are randomly selected from each category. These samples are then combined with data from other tasks to form a new memory \(R_n\). As training progresses, the memory is continuously updated by adding data from the most recent task and replacing some of the existing data from other tasks at a certain ratio. The new memory \(R_n\) is defined as:

\begin{equation}
R_n = \bigcup_{i=1}^{n} \left\{ \text{sample}(D_{c,i}, S_{c,i}) \mid c \in C_i \right\},
\end{equation}
where \( C_i \) is the set of categories in task \( i \), and the function \( \text{sample}(D_{c,i}, S_{c,i}) \) refers to randomly selecting \( S_{c,i} \) samples from category \( c \) in task \( i \)'s dataset \( D_{c,i} \).

When dealing with complex data for autonomous driving, memory replay, while mitigating catastrophic model forgetting, still has limitations in ensuring robust generalization across diverse scenarios. Considering the importance of detailed information in autonomous driving models, the introduction of knowledge distillation via distillation can more effectively convey essential information and promote smoother transitions between tasks. This enhances the model's ability to generalize across varying conditions and improves its overall performance. Along with the memory replay strategy, we also integrate a knowledge distillation approach to improve the retention of prior knowledge, ensuring a more stable and consistent learning process.

\noindent\textbf{Knowledge Distillation.} In our approach, the previous task model $M_t$ serves as the “teacher” and the current model $M_s$ as the “student.” Rather than distilling all tokens equally, we only apply soft‐label distillation to those tokens for which the teacher is sufficiently confident; for less certain tokens we rely solely on ground‐truth supervision. Given the teacher’s logits $z^T_{i,t}$ for sample $i$ at token position $t$, we first compute temperature‐scaled probabilities. Here, $z^T_{i,t}\in\mathbb{R}^{K}$ denotes the $K$-dimensional logit vector over the vocabulary.
\begin{equation}
  p^T_{i,t} = \operatorname{softmax}\bigl(z^T_{i,t}/T\bigr)\,,
\end{equation}
where $T$ is the distillation temperature. We then measure the token‐level confidence by
\begin{equation}
  c_{i,t} = \max_k p^T_{i,t}[k]\,,
\end{equation}

and map it to a per‐token distillation weight
\begin{equation}
  \alpha_{i,t} = 
  \begin{cases}
    0, & c_{i,t} < \tau,\\[6pt]
    \displaystyle\frac{c_{i,t}-\tau}{1-\tau}\,\alpha_{\max}, 
      & c_{i,t} \ge \tau,
  \end{cases}
\end{equation}
where $\tau$ is the confidence threshold and $\alpha_{\max}$  the maximum distillation ratio.

Similarly, the student’s temperature‐scaled probabilities are
\begin{equation}
  p^S_{i,t} = \operatorname{softmax}\bigl(z^S_{i,t}/T\bigr)\,.
\end{equation}
For each valid (non‐padding) token we compute the standard cross‐entropy loss $\mathcal{L}_{GT}^{(i,t)}$ using the ground‐truth label, and the KL‐divergence

\begin{equation}
  \mathcal{L}_{KD}^{(i,t)} 
  = \sum_k p^T_{i,t}[k]\,
    \log\frac{p^T_{i,t}[k]}{p^S_{i,t}[k]}
  \times T^2
\end{equation}
for soft‐label alignment. The overall loss for a batch is then
\begin{equation}
  \mathcal L_\text{replay}
    = \frac{1}{\sum_{i,t} m_{i,t}}
    \sum_{i,t} m_{i,t}\Bigl[
      (1-\alpha_{i,t})\,\mathcal{L}_{GT}^{(i,t)}
      + \alpha_{i,t}\,\mathcal{L}_{KD}^{(i,t)}
    \Bigr],
\end{equation}
where $m_{i,t}\in\{0,1\}$ masks out padding tokens. This selective distillation ensures that the student inherits only the teacher’s highly reliable “dark knowledge,” while avoiding the teacher’s uncertain or incorrect predictions, thus striking a better balance between retaining past‐task performance and adapting to new tasks.

\subsection{Embedding Projection}

In the continual learning setting of autonomous driving, memory replay and knowledge distillation help the model learn abstract features and decision logic. However, their effectiveness depends on the teacher model's accuracy and generalization over replayed data. Moreover, these methods focus on output similarity, rather than directly preserving feature continuity across tasks. To mitigate the bias caused by training on new tasks, we introduce a model regularization phase during each task's learning process. In the context of large-scale VLMs for autonomous driving, traditional parameter regularization and gradient projection methods used in smaller networks are not directly applicable. Thus, we propose incorporating a feature projection layer into the model's embedding block (Figure~\ref{Pro}). By updating this projection and applying feature-level regularization during training, we maintain the structural consistency of the model’s representations across tasks.

In our work, each task is associated with a dedicated projection layer \(P_n\), designed to transform the merged feature embeddings from multimodal inputs (text and image) into a task-specific feature space \(F_n\). Each projection layer \(P_n\) consists of a weight matrix \(\mathbf{W}_n\), which linearly transforms the feature vector from the preceding layer into a new task-specific feature space:
\begin{equation}
    \mathbf{F}_n = \mathbf{W}_n \mathbf{e},
\end{equation}
where \(\mathbf{e}\) represents the feature embeddings obtained from the multimodal input processing. The transformation is designed to ensure that each task’s feature representation is both unique and aligned with the task-specific requirements. It is important to note that the projection layers do not freeze or isolate the underlying embedding $\mathbf{e}$. Instead, each task learns its own projection $P_n$ that maps $\mathbf{e}$ into a lower-dimensional, task-specific space. This design serves two purposes. First, by constraining the similarity of the projected features rather than the raw embeddings, we reduce the variance introduced by task-dependent input distributions, making the feature space more stable across tasks. Second, even if $\mathbf{e}$ evolves during training, the projection spaces act as an anchoring mechanism: the task-specific representations  Eq.~(9) remain aligned through the regularization, effectively controlling representational drift without over-constraining the backbone VLM.

\begin{algorithm}
\caption{continual learning for AD VLM}
\label{alg:cl_vlm}

\textbf{Input}\\
\noindent \hangindent=3em \hangafter=1
Multi-task datasets \(\{D_1, \ldots, D_n\}\) with tasks \(T_n\), VLM model \(\phi\) with pre-trained components (ViT, T5), memory buffer \(R\)

\textbf{Output}\\
\noindent \hangindent=3em \hangafter=1
Trained VLM model \(\phi\) 

\begin{algorithmic}[1]
\STATE Initialize \(\phi\) and \(R\)
\FOR{\(n = 1\) to \(N\)}
    \STATE Load task-specific dataset \(D_n\) and task \(T_n\)
    \IF{\(n == 1\)}
        \STATE Train \(\phi\) on \(D_1\)
        \STATE Populate \(R\) using TF-IDF and K-means on \(D_1\)
    \ELSE
        \FOR{each iteration}
            \STATE Sample \(x\) from \(D_n \cup R\)
            \IF{\(x \in D_n\)}
                \STATE Update \(\phi\) using projection layer \(P_n\)
            \ELSE
                \STATE Replay using model \(M_t\) from \(T_{n-1}\)
            \ENDIF
        \ENDFOR
        \STATE Refresh \(R\) with samples from \(D_n\)
    \ENDIF
    \STATE Evaluate \(\phi\) on \(T_n\)
\ENDFOR
\end{algorithmic}
\end{algorithm}

During training, we employ a regularization approach that involves computing the mean squared error (MSE) between the feature outputs of the projection layers for the current and previous tasks. This MSE is used to formulate a regularization loss $\mathcal{L}_{pro}$, which serves to minimize the drift in feature space across tasks, thereby mitigating catastrophic forgetting, as the following:
\begin{equation}
    \mathcal L_{pro}(n, m) = \frac{1}{d} \sum_{i=1}^d \left(\mathbf{F}_n[i] - \mathbf{F}_m[i]\right)^2,
\end{equation}
where \(d\) is the dimensionality of the feature vectors, and \(i\) indexes the elements of these vectors, \(n\) represents the current task, \(m\) represents a previous task.

As the model encounters new tasks, we dynamically adapt the network by adding new projection layers while freezing the parameters of previous layers. This ensures that the newly learned tasks do not disrupt the knowledge encoded in the weights of earlier tasks. The overall regularization loss for the network, accumulated over all tasks up to the current one, is given by:
\begin{equation}
    \mathcal L_{pro}(n) = \sum_{m=1}^{n-1} \mathcal L_{pro}(n, m),
\end{equation}
thus maintaining a continuity of learning without catastrophic interference.

The training procedure optimizes the combined loss, which includes task-specific losses and the projection layer regularization loss. The loss function for the network is formulated as:
\begin{equation}
    \mathcal Loss = \mathcal{L}_{GT} + \lambda \mathcal{L}_{pro},
\end{equation}

\noindent where \(\mathcal{L}_{pro}\) represents the projection layer regularization loss and \(\lambda\) is a hyperparameter that balances the importance of the regularization term relative to the task-specific loss.

This approach ensures that our model adapts to new tasks while retaining critical information from previous tasks, thereby facilitating effective continual learning in a complex, multi-task environment. The algorithm for this approach will be presented in the supplementary materials. The total loss function can be formulated as the following:
\begin{equation}
\mathcal Loss_{total} = 
\begin{cases} 
\alpha \mathcal  L_{\text{replay}}  & \text{if replay data}, \\
\lambda \mathcal L_{{pro}} + \mathcal L_{\text{GT}} & \text{otherwise}.
\end{cases}
\end{equation}
The entire continual learning workflow is summarized in Algorithm~\ref{alg:cl_vlm}.

\section{Experiments}\label{sec4}
\subsection{Experimental Settings}

\begin{table*}[!t]
\centering
\fontsize{8pt}{8pt}\selectfont

\renewcommand{\arraystretch}{1}
\resizebox{1\textwidth}{!}{%
\begin{tabular}{lcccccccc}
\toprule
\textbf{Method} &
  \multicolumn{1}{c}{BLEU-1} &
  \multicolumn{1}{c}{BLEU-2} &
  \multicolumn{1}{c}{BLEU-3} &
  \multicolumn{1}{c}{BLEU-4} &
  \multicolumn{1}{c}{METEOR} &
  \multicolumn{1}{c}{ROUGE\_L} &
  \multicolumn{1}{c}{CIDEr}\\ 
    \cmidrule{1-8}

Joint   & 66.82 & 60.34 & 54.74 & 49.58 & 36.07 & 72.46 & 3.2 \\
Vanilla  & 36.67 & 31.10 & 28.06 & 25.75 & 21.13 & 53.29 & 1.8 \\
DriveLM-Agent  & - & - & - & 53.09  & 36.19   & 65.58 & 2.8 \\
    \cmidrule{2-8}
LwF   & 44.45 & 39.75 & 36.63 & 34.08 & 24.75 & 56.02 & 2.1 \\
EWC   & 46.51 & 42.47 & 39.54 & 39.67 & 27.40 & 63.10 & 2.4 \\
DER    & 65.27 & 60.74 & 56.94 & 53.29 & 38.83 & 71.44 & 3.1 \\
\textbf{Ours}  &\textbf{71.25} & \textbf{66.26} & \textbf{61.84} & \textbf{57.62} & \textbf{41.24} & \textbf{75.84} & \textbf{3.4} \\
\bottomrule
\end{tabular}%
}
\caption{Comparison of different methods on various metrics, demonstrating the effectiveness of our proposed continual learning with VLM model for autonomous driving across multiple performance indicators. }
\label{Experiment Result}
\end{table*}
\textbf{Dataset.}
We use the DriveLM~\cite{drivelm} dataset, a multimodal dataset designed for autonomous driving research, containing large amounts of image and text data across various conditions—including sunny, rainy, and nighttime scenarios—to support the development of vision‑language models.In our experiments, we divided the DriveLM dataset into 4 tasks: Perception, Prediction, Planning, and Behavior, and each task contains a corresponding training/validation/testing set. More details please refer to supplementary documents.


\noindent\textbf{Metrics.}~In our experiment, we adopt a variety of evaluation metrics to comprehensively assess model performance. Specifically, BLEU\cite{Bleu}, METEOR\cite{Meteor}, ROUGE\cite{Rouge}, and CIDEr\cite{Cider} are used following the official DriveLM evaluation protocol. Although these metrics originate from natural-language generation, DriveLM assess task correctness rather than linguistic fluency. This is because the dataset expresses multiple autonomous-driving subtasks in QA form: perception questions correspond to object detection and attribute recognition; prediction questions describe the future motion of surrounding agents; planning questions encode rule-based driving decisions; and behavior questions reflect high-level driving intent. Thus, the textual answers function as structured labels for these subtasks, and the metric scores reflect how accurately the model solves Perception, Prediction, Planning, and Behavior tasks. 

Under our continual learning setup, after training each task $T_n$, the model is re-evaluated on all previously learned tasks. As a result, questions from early tasks may accumulate multiple predictions across training stages. In line with standard continual-learning practice, after completing all tasks, we report the scores of the final model on the test sets of every task.


\subsection{Analysis}

\noindent\textbf{Compared Methods.}
We compare our method with the following methods on the DriveLM dataset~\cite{drivelm}: the weight regularization method EWC~\cite{EWC}, the function regularization method LwF~\cite{LwF}, and the memory replay method DER~\cite{DER} to validate the effectiveness of our approach. In addition, we provide a lower bound (Vanilla), which simply performs gradient updating without any countermeasures against forgetting, and a relative upper bound (Joint), which trains all tasks jointly. Moreover, we compare another method that uses a joint training approach but is based on a different model, referred to as DriveLM-Agent~\cite{drivelm}, to demonstrate the performance differences in handling multitask learning across various model architectures.

\noindent\textbf{Implementation Details.} Our implementation is based on the EM‑VLM4AD framework~\cite{EM-VLM4AD}, utilizing a Vision Transformer (ViT)\cite{ViT} for multi-view image embedding and a pre-trained T5 model for text processing, enabling Visual Question Answering (VQA) in autonomous driving. Our EM‑VLM4AD (T5‑Base) model is lightweight, with 235M parameters, 9.47B FLOPs, and 0.94GB memory, achieving under 1s latency per sample—suitable for near real‑time use. 
We implement our model in PyTorch\cite{pytorch} and train all tasks on a single NVIDIA L20 GPU, with each task trained for 4 epochs. In our four-task setting, the complete continual learning pipeline takes about 20 hours to finish, and the additional overhead introduced by replay (small batch fraction), knowledge distillation (one extra teacher forward per replay batch), and projection layers (\(<\)1M parameters, \(<\)0.5\% FLOPs) remains lightweight relative to the VLM backbone. For hyper‑parameters, we use a learning rate of 1e‑4, a weight decay of 0.05, a batch size of 4, a temperature parameter set to 2 and an $\alpha$ value of 0.7 for the knowledge distillation part. A dynamic weight parameter adjusts the projection regularization loss, where the $\lambda$ parameter is initialized to 0.05 and halved progressively with each task. More details of experimental hyperparameter settings and experiments refer to supplementary materials.

\begin{table}
    \resizebox{\linewidth}{!}{
    \begin{tabular}{ccc|cccc}
    \toprule
    \multicolumn{1}{c}{ER} &
    \multicolumn{1}{c}{KD} &
    \multicolumn{1}{c|}{PRO} &
      \multicolumn{1}{c}{BLEU-4} &
      \multicolumn{1}{c}{METEOR} &
      \multicolumn{1}{c}{ROUGE\_L} &
      \multicolumn{1}{c}{CIDEr}\\ 
        \cmidrule{1-7}
    \checkmark  &  &  & 55.80 & 39.89 & 75.03 & 3.20 \\
    \checkmark  & \checkmark  &  & 56.03 & 40.21 & 75.20 & 3.15 \\
     &  & \checkmark &31.80 & 23.94 & 53.19 & 1.85 \\
     \checkmark&   & \checkmark & 56.42 & 40.44 & 75.30 & 3.21 \\
    \checkmark& \checkmark & \checkmark &   \textbf{57.62} & \textbf{41.24} & \textbf{75.84} & \textbf{3.36} \\
    \bottomrule
    \end{tabular}
    } 
    \caption{Ablation studies on the effects of individual components and their combinations in methods.}
    \label{ablation study} 

\end{table}

\subsection{Quantitative Results}
 
As shown in Table~\ref{Experiment Result}, our model significantly outperforms existing continual learning methods across all key metrics, achieving the highest BLEU, METEOR, ROUGE, and CIDEr scores. Specifically, it reaches a BLEU-1 score of $71.25\%$, surpassing the next best method (Joint) by $4.43\%$, and improves BLEU-4 and METEOR by $8.04\%$ and $5.17\%$, respectively, demonstrating stronger capability for nuanced, context-aware language generation in autonomous driving. The CIDEr score improvement from 3.2 to 3.4 further confirms the model’s ability to produce relevant, human-aligned image descriptions.
Our model also outperforms EWC, LwF, and DER, exceeding them by $17.95\%$, $23.54\%$, and $4.33\%$ in BLEU-4, respectively. These results highlight the limitations of regularization approaches and the benefits of combining memory replay with advanced knowledge distillation and projection-layer regularization, which enable better task retention and multimodal understanding. Moreover, our CIDEr score of 3.4, the highest among all methods, demonstrates the effectiveness of dynamic projection regularization in preventing feature drift and maintaining robustness across driving scenarios.

\subsection{Ablation Studies}

In order to analyze and study the effects of the components of our method, we designed several ablation experiments, with results shown in Table~\ref{ablation study}.

\noindent\textbf{Memory replay and Knowledge Distillation.} Using memory replay/experience replay (ER) alone significantly enhances the model's continual learning performance. While knowledge distillation (KD) has a limited effect on traditional metrics, it effectively improves the CIDEr score by smoothing probability distributions, thereby reducing overconfident predictions and enhancing prediction stability. 

\noindent\textbf{Projection regularization (Pro).} Our proposed Pro plays a vital role by constraining the feature space, helping the model retain essential task-specific information and improve adaptability across tasks. The last line in the table demonstrates that combining all components yields strong performance across all variants, proving the overall effectiveness of our continual learning method. Further experimental validation of KD’s role in maintaining task-specific focus and Pro’s contribution to enhanced model adaptability and performance is provided in the supplementary materials.

\begin{table}
\renewcommand{\arraystretch}{1.0}
\resizebox{\columnwidth}{!}{
\begin{tabular}{ccccc}
\toprule
\multicolumn{1}{c|}{Method} &
\multicolumn{1}{c}{Exist} &
  \multicolumn{1}{c|}{Comparison} &
  \multicolumn{1}{c}{Average $\uparrow$}&
  \multicolumn{1}{c}{Forget $\downarrow$ } \\ 
    \cmidrule{1-5}
w/o CL  & 70.47  & 62.56 & 40.78 & 14.40\% \\
w/ CL  & \textbf{77.49}  & \textbf{67.54} & \textbf{49.27}  & \textbf{4.5\%} \\
\bottomrule
\end{tabular}
}
\caption{Performance of our method on Nuscenes-QA Dataset.}
\label{NuscenesQA}

\end{table}

\begin{table}
\renewcommand{\arraystretch}{1.0}
\resizebox{\columnwidth}{!}{
\begin{tabular}{c|cccc}
\toprule

\multicolumn{1}{c|}{Method} &
  \multicolumn{1}{c}{BLEU-4} &
\multicolumn{1}{c}{METEOR} &
  \multicolumn{1}{c}{ROUGE\_L} &
  \multicolumn{1}{c}{CIDEr}\\
    \cmidrule{1-5}

w/o CL  &  38.74& 28.72 & 61.40 & 2.37 \\
w/CL  &  \textbf{54.05} & \textbf{38.87} & \textbf{73.28} & \textbf{3.17} \\
\bottomrule
\end{tabular}
}
\caption{Performance Comparison of LLaMA-Adapter-v2 Backbone with and without Continual Learning (CL). }
\label{llama_backbone}
\end{table}

\begin{figure*}[!t]
    \centering
    \includegraphics[width=0.8\linewidth,height=0.95\linewidth]{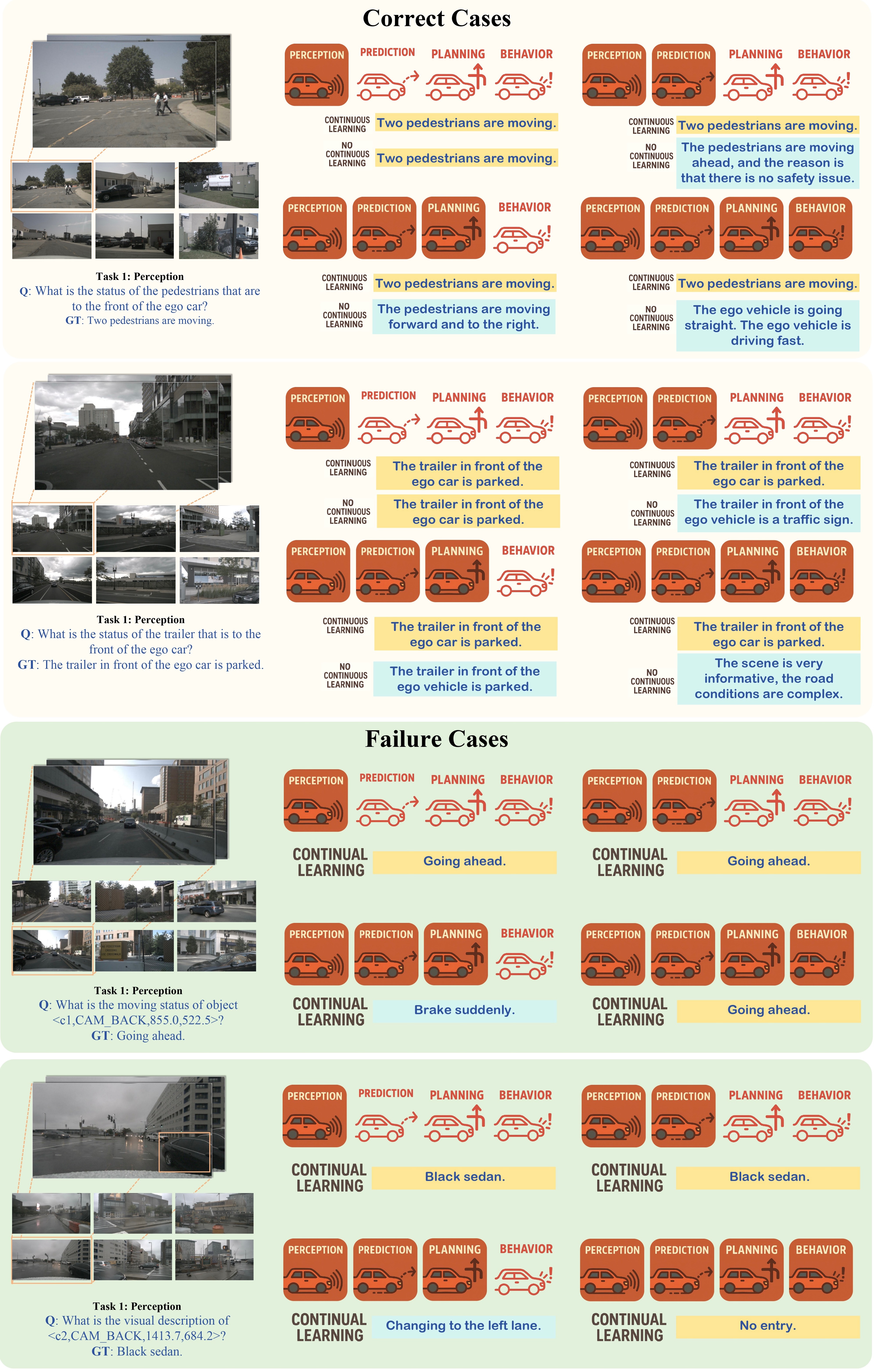}
    \caption{Examples of correct and failure answer generations from our model.}
    \label{cases}
\end{figure*}
\noindent\textbf{Performance of Our Method on Other VQA Datasets.}
The table~\ref{NuscenesQA} illustrates the effectiveness of our proposed method on the Nuscenes-QA~\cite{nuscenesqa} dataset, demonstrating significant improvements across critical evaluation metrics. With the incorporation of continual learning (CL), our approach achieves notably higher accuracy in both the 'Exist' and 'Comparison' tasks, accompanied by a substantial increase in the overall 'Average' score. Additionally, the marked reduction in the 'Forget' rate underscores the model's enhanced capability for knowledge retention and adaptability in evolving environments.

\noindent\textbf{Adaptability of CL method with Other VLM model.} 

To evaluate the adaptability of our continual learning (CL) approach, we replaced the backbone in our VLM framework with LLaMA-Adapter-v2~\cite{llama}. As shown in Table~\ref{llama_backbone}, applying CL ("w/ CL") improves BLEU, METEOR, ROUGE\_L, and CIDEr scores over the non-CL baseline ("w/o CL"). These gains reflect enhanced fluency, contextual accuracy, and alignment with human references in image-grounded tasks. This result demonstrates the robustness of our CL method across architectures and highlights its flexibility for integration into diverse vision-language frameworks, especially in autonomous driving scenarios.

\subsection{Case Analysis}

Figure~\ref{cases} presents examples of correct and failure cases in our multitask autonomous driving framework. Note that each question in the figure is evaluated at four different stages of continual learning. After completing task $T_n$, the model is re-evaluated on all previously learned tasks; therefore, a single perception question yields four answers corresponding to the checkpoints obtained after training Perception, Prediction, Planning, and Behavior. This setup allows us to visualize whether the model preserves earlier knowledge as new tasks are introduced.

In the correct examples, with joint training on perception, prediction, planning, and behavior tasks—and the use of continual learning methods—the model maintains its accuracy on the perception task. For instance, in the pedestrian scenario, it consistently answers “Two pedestrians are moving,” and in the trailer scenario, it consistently responds “The trailer in front of the ego car is parked,” across perception, prediction, planning, and behavior stages. By contrast, without continual learning, the model’s outputs become dominated by information from later tasks, causing crucial details from earlier perception subtasks to be lost and illustrating catastrophic forgetting in a multitask setting.

Despite continual learning, data imbalances still cause errors. For example, in the behavior task, scarce parked‑vehicle samples lead the model to predict “Going ahead” when the vehicle has braked, or to mislabel a black sedan as “Changing to the left lane” or “No entry” in trailer scenarios. These failure cases underscore the challenge of rare samples—see the supplementary materials for further analysis.

\section{Conclusion}\label{sec5}
In this paper, we presented a continual learning approach applied to vision language models (VLMs) for autonomous driving with VQA task. The method combines memory replay and knowledge distillation, and demonstrates its strong performance on the DriveLM dataset by introducing a projection layer for regularization in feature embedding. Extensive experimental results demonstrate the effectiveness and generalizability of the proposed model. In future work, we will explore more dynamic continual learning strategies to further improve model robustness.

{
    \small
    \bibliographystyle{ieeenat_fullname}
    \bibliography{main}
}

\input{X_suppl}

\end{document}

%% file: preamble.tex
\usepackage[dvipsnames]{xcolor}


%% file: X_suppl.tex
\clearpage
\setcounter{page}{1}
\maketitlesupplementary

\section{Hyperparameters}
\label{sec:hyperparameters}

\subsection{Memory Size} Table \ref{Memory Size table} shows how the model's performance varies with different sizes of replay data. Overall, memory replay improves the overall continual learning performance of the model regardless of the size of the replay data. Based on the data in the table, it can be seen that the model improves with larger replay data in the two metrics BLUE and METEOR, because BLUE and METEOR focus more on text matching. However, although a large amount of replay data can reduce forgetting, it poses a significant challenge to the model's memory as the replay data becomes larger. Also, according to the ROUGE\_L and CIDEr metrics, it can also be seen that an increase in the amount of model replay data has a positive effect on the model's text-image fusion understanding, but too much replay data can also limit the model's learning of new tasks, which reduces the overall continual learning performance.

\subsection{Temperature Setting}
Table \ref{Temperature Setting table} shows the role of the knowledge distillation model for model Memory replay performance for different temperature and parameter settings. For the same temperature parameter T setting, different $\alpha$ values indicate different compositions of model loss when memory replay is subjected to knowledge distillation. As $\alpha$ increases, increasing the weight of knowledge distillation loss means that the model relies more on the knowledge of the teacher model, which may help the student model learn some intrinsic patterns and relationships that are not easily observed at the simple labeling level. On the other hand, for different temperature coefficients, higher T values may result in a weaker gradient signal being passed from the teacher model to the student model. In this case, the backpropagated signals received by the student model during training may not be sufficient to effectively tune its parameters, resulting in a model that is not as effective at learning past knowledge. Therefore, in our model, we chose T = 2 and $\alpha$ = 0.7, which is the parameter of function 4 in the main body, as the model parameters for knowledge distillation.

\subsection{Lambda Setting}
Table~\ref{lambda setting table} demonstrates the usefulness of the projection regularization approach for continual learning of the model for different settings of the projective regularization coefficients $\lambda$, which is the parameter of function 8 in the main body. The table shows that setting larger or smaller fixed regularization coefficients does not work for the overall architecture of the model, which changes as tasks increase. In our approach, the model projection layers are incremented as tasks increase, and each task projection layer constrains the model by calculating the distance to the previous task projection layer. With fixed $\lambda$ coefficients, the model regularization data increases as the model task increases, which has a large impact on model migration and stability, and thus model performance is poor. The dynamically decreasing $\lambda$ with the increase of model tasks can make the multi-projection layer gap in different tasks are maintained at a relatively stable amount, so that the model can remain stable during the migration process between tasks.

\begin{table}
\renewcommand{\arraystretch}{1.1}
\resizebox{\columnwidth}{!}{
\begin{tabular}{c|cccc}
\toprule

\multicolumn{1}{c|}{Memory Size} &
  \multicolumn{1}{c}{BLUE-4} &
  \multicolumn{1}{c}{METEOR} &
  \multicolumn{1}{c}{ROUGE\_L} &
  \multicolumn{1}{c}{CIDEr}\\ 
    \cmidrule{1-5}
1000  & 52.00 & 37.11 & 71.32 & 3.06 \\
3000  & 53.07 & 37.79 & 72.05 & 3.12 \\
5000  & \textbf{55.33} & \textbf{40.17} & \textbf{74.69} & \textbf{3.41} \\
8000  &56.52 & 41.10 & 73.56 & 3.35 \\
\bottomrule
\end{tabular}
}
\caption{The comparison of different memory sizes in the model on various metrics demonstrates the effect of different memory sizes on the continual learning approach.}
\label{Memory Size table}
\end{table}

\begin{table}
\renewcommand{\arraystretch}{1.1}
\resizebox{\columnwidth}{!}{
\begin{tabular}{c|cccc}
\toprule

\multicolumn{1}{c|}{Temperature Setting} &
  \multicolumn{1}{c}{BLUE-4} &
  \multicolumn{1}{c}{METEOR} &
  \multicolumn{1}{c}{ROUGE\_L} &
  \multicolumn{1}{c}{CIDEr}\\ 
    \cmidrule{1-5}

T=2/$\alpha$=0.6  & 34.08 & 24.74 & 56.02 & 2.04 \\
T=2/$\alpha$=0.7  & \textbf{55.33} & \textbf{40.17} & \textbf{74.69} & \textbf{3.41} \\
T=3/$\alpha$=0.7  & 41.37 & 31.20 & 54.47 & 2.64 \\
\bottomrule
\end{tabular}
}
\caption{The comparison of different knowledge distillation parameters in the model on various metrics demonstrates the effect of different temperature coefficients and parameters for the continual learning approach.}
\label{Temperature Setting table}
\end{table}

\begin{table}
\renewcommand{\arraystretch}{1.1}
\resizebox{\columnwidth}{!}{
\begin{tabular}{c|cccc}
\toprule

\multicolumn{1}{c|}{lambda setting} &
  \multicolumn{1}{c}{BLUE-4} &
  \multicolumn{1}{c}{METEOR} &
  \multicolumn{1}{c}{ROUGE\_L} &
  \multicolumn{1}{c}{CIDEr}\\ 
    \cmidrule{1-5}

$\lambda$=0.01  & 53.41 & 38.42 & 72.43 & 3.01 \\
$\lambda$ change  & \textbf{55.33} & \textbf{40.17} & \textbf{74.69} & \textbf{3.41} \\
$\lambda$=0.1  &55.98 & 40.16 & 74.2 & 3.11 \\
\bottomrule
\end{tabular}
}
\caption{The comparison of different projection regularization coefficients in the model on various metrics demonstrates the effect of different regularization coefficients in for the continual learning approach.}
\label{lambda setting table}
\end{table}

\begin{figure*}[htbp]
  \centering
  \begin{subfigure}{.33\textwidth}
    \centering
    \includegraphics[width=\linewidth]{ 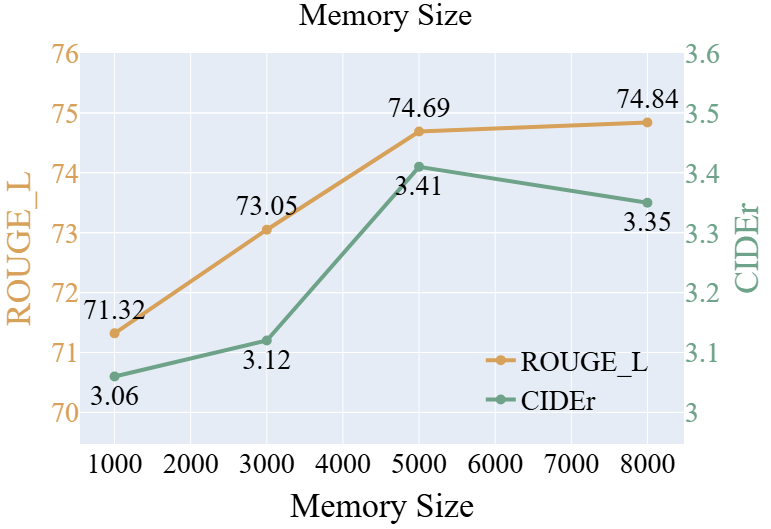}  
    \caption{Impact of Memory Size Settings}
    \label{fig Memory size}
  \end{subfigure}\hfill
  \begin{subfigure}{.33\textwidth}
    \centering
    \includegraphics[width=\linewidth]{ 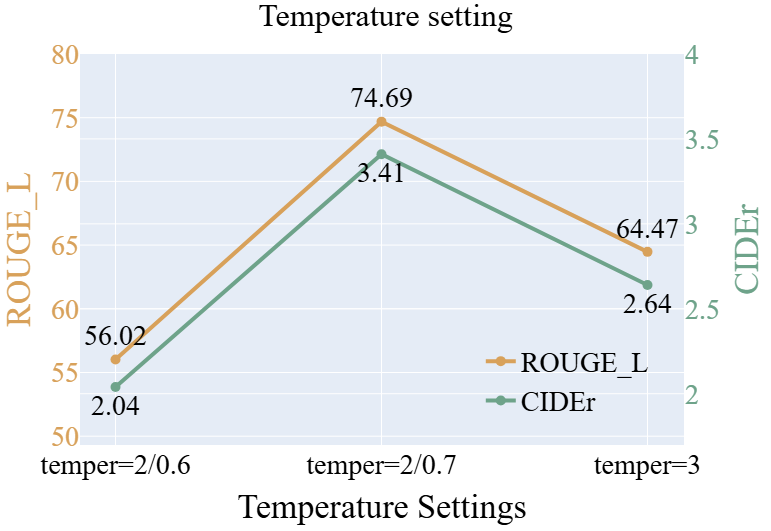}  
    \caption{Impact of Temperature Parameter Settings}
    \label{fig temperature setting}
  \end{subfigure}\hfill
  \begin{subfigure}{.33\textwidth}
    \centering
    \includegraphics[width=\linewidth]{ 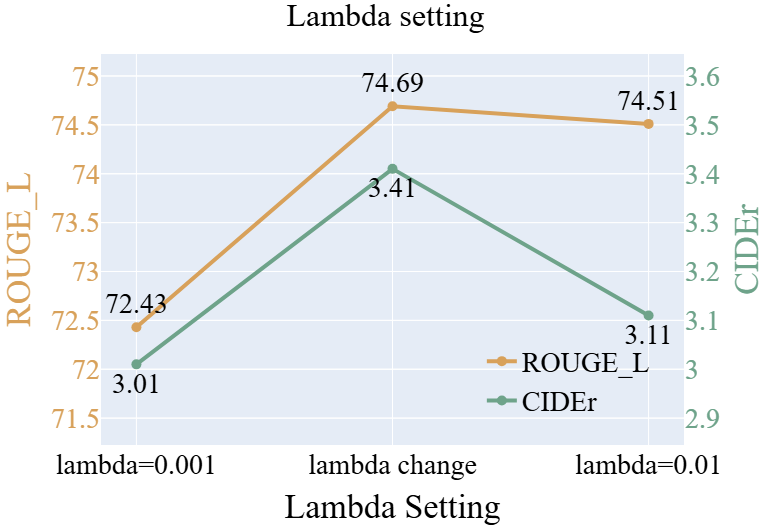}  
    \caption{Impact of Lambda Values Settings}
    \label{fig lambda setting}
  \end{subfigure}\hfill
  \caption{Illustration of a detailed visual analysis about how different configuration settings impact the performance metrics of our continual learning model for autonomous driving. Each subfigure presents a unique aspect of model tuning and its direct correlation with the performance in language and vision tasks. }
\end{figure*}

\section{Knowledge Distillation and Projection Regularization}

\subsection{Knowledge Distillation}
Knowledge Distillation (KD) is introduced in the model to encourage smoother and more balanced output probability distributions. This is particularly critical in autonomous driving tasks, where stability and reliability of predictions directly influence performance in real-world scenarios. To analyze this effect, we examine the entropy of the predicted probability distributions across different tasks at the final checkpoint, which corresponds to the completion of training on all tasks. The entropy $H(p)$ of a probability distribution $p$ is defined as:
\begin{equation}
H(p) = - \sum_{i=1}^C p_i \log(p_i),
\end{equation}
where $C$ is the number of classes, and $p_i$ represents the predicted probability for class $i$. Higher entropy indicates smoother predictions, avoiding extreme confidence in a single class.

\begin{table}
\renewcommand{\arraystretch}{1.2}
\resizebox{\columnwidth}{!}{
\begin{tabular}{c|cc|ccc}
\toprule

\multicolumn{1}{c|}{Task} &
\multicolumn{1}{c}{ER} &
\multicolumn{1}{c|}{ER+KD} &
\multicolumn{1}{c}{Test loss w/Pro} &
\multicolumn{1}{c}{Test loss w/o Pro} &
\multicolumn{1}{c}{Difference} \\
\cmidrule{1-6}

Perception  & 0.6153	& 0.6369 &0.1849	&0.1848	&0.0001\\
Prediction & 0.3515	    & 0.2964 & 0.1212	&0.1999	&-0.0787\\
Planning  &0.6120	    & 0.6301 &0.1084	&0.1111	&-0.0027 \\
Behavior  &0.7019	    &0.7856 & 0.1024	&0.1052	&-0.0028\\
\bottomrule
\end{tabular}
}
\caption{Comparison of average entropy and test loss across tasks. The table shows entropy with Experience Replay (ER) alone and ER with Knowledge Distillation (KD), as well as test loss with and without Projection Regularization (Pro).}
\label{kdpro}
\end{table}

Table~\ref{kdpro} shows the average entropy for tasks under two settings: using Experience Replay (ER) alone and combining ER with KD. 
For tasks Perception, Planning, and Behavior, the addition of KD increases entropy values, indicating a smoother distribution of predicted probabilities. In Behavior, for example, entropy rises from 0.7019 to 0.7856. This suggests that KD mitigates overconfident predictions, enhancing robustness in complex environments where the model must avoid excessive certainty in a single outcome.

In contrast, for Prediction, entropy slightly decreases from 0.3515 to 0.2964. This reflects a beneficial property of KD in tasks with limited plausible outcomes, where the model is guided to focus on relevant classes, improving prediction certainty while maintaining reliability. These findings underline the versatility of KD in balancing prediction smoothness and confidence depending on task-specific requirements.

\subsection{Projection Regularization}
Projection Regularization (Pro) imposes constraints on the feature space, which helps prevent overfitting and enhances the model's generalization to unseen data. To evaluate its effectiveness, we analyze the test loss across various tasks with and without Pro. The test loss $\mathcal L_{\text{test}}$ is calculated as:
\begin{equation}
\mathcal L_{\text{test}} = - \frac{1}{N} \sum_{j=1}^N \sum_{i=1}^C y_{j,i} \log \hat{y}_{j,i},
\end{equation}
where $y_{j,i}$ and $\hat{y}_{j,i}$ denote the ground truth and predicted probabilities for sample $j$ and class $i$, respectively.

Table~\ref{kdpro} summarizes the test loss values under both settings. The results demonstrate that Pro consistently reduces test loss across most tasks, highlighting its role in improving generalization. For the Prediction task, test loss significantly decreases from 0.1999 to 0.1212, indicating that Pro effectively constrains feature dependencies, enabling the model to adapt better to unseen scenarios. While the reductions in Planning and Behavior tasks are smaller, they still reflect the robustness that Pro introduces. 

For the Perception task, the change in test loss is negligible, suggesting that Pro neither harms nor substantially impacts performance in simpler scenarios. Overall, these results validate the effectiveness of Pro in promoting adaptability and robustness in real-world applications.

Both KD and Pro contribute critical enhancements to the model's performance in autonomous driving tasks. KD ensures smoother and more stable predictions, reducing overconfidence in complex environments while maintaining appropriate certainty in constrained tasks. Pro, on the other hand, strengthens generalization by constraining the feature space, improving test performance across diverse tasks. These complementary mechanisms play a vital role in ensuring the reliability and robustness of the model in safety-critical applications.

\section{Algorithm}
\begin{algorithm}
\caption{continual learning for AD VLM}

\textbf{Input}\\
\noindent \hangindent=3em \hangafter=1
Multi-task datasets \(\{D_1, \ldots, D_n\}\) with tasks \(T_n\), VLM model \(\phi\) with pre-trained components (ViT, T5), memory buffer \(R\)

\textbf{Output}\\
\noindent \hangindent=3em \hangafter=1
Trained VLM model \(\phi\) 

\begin{algorithmic}[1]
\STATE Initialize \(\phi\) and \(R\)
\FOR{\(n = 1\) to \(N\)}
    \STATE Load task-specific dataset \(D_n\) and task \(T_n\)
    \IF{\(n == 1\)}
        \STATE Train \(\phi\) on \(D_1\)
        \STATE Populate \(R\) using TF-IDF and K-means on \(D_1\)
    \ELSE
        \FOR{each iteration}
            \STATE Sample \(x\) from \(D_n \cup R\)
            \IF{\(x \in D_n\)}
                \STATE Update \(\phi\) using projection layer \(P_n\)
            \ELSE
                \STATE Replay using model \(M_t\) from \(T_{n-1}\)
            \ENDIF
        \ENDFOR
        \STATE Refresh \(R\) with samples from \(D_n\)
    \ENDIF
    \STATE Evaluate \(\phi\) on \(T_n\)
\ENDFOR
\end{algorithmic}
\end{algorithm}

The proposed algorithm integrates continual learning strategies for Vision-Language Models (VLMs) in autonomous driving, aiming to mitigate catastrophic forgetting while ensuring adaptability to new tasks. The algorithm operates iteratively across multiple tasks \( T_n \), leveraging pre-trained components 7and a dynamically updated memory buffer \( R \). The key steps are described below:

\begin{itemize}
    \item \textbf{Initialization:} The VLM model \( \phi \) and memory buffer \( R \) are initialized to prepare for training across tasks.
    \item \textbf{Task-Specific Training:}
For each task \( T_n \), the corresponding dataset \( D_n \) is loaded.
For the first task \( T_1 \), the model is trained directly on \( D_1 \), and the memory buffer \( R \) is populated using TF-IDF and K-means clustering to ensure representativeness and diversity of stored samples.
    \item \textbf{Memory Replay and Regularization:}
For tasks \( T_n \) (\( n > 1 \)), the model alternates between training on current task data \( D_n \) and replaying samples from the memory \( R \).
During training, task-specific projection layers \( P_n \) are updated to transform feature embeddings into task-specific spaces. This regularization helps maintain feature alignment across tasks, mitigating interference.
    \item \textbf{Dynamic Memory Update:}
After training on each task, the memory buffer \( R \) is refreshed by sampling representative data from both the current task and past tasks. This ensures the memory remains balanced and relevant for continual learning.
    \item \textbf{Evaluation:} Once trained on \( T_n \), the model is evaluated to measure its performance on the current task while retaining knowledge from previous tasks.
\end{itemize}

By combining memory replay, knowledge distillation, and embedding projection, the algorithm effectively balances learning new tasks and preserving prior knowledge, making it suitable for complex and dynamic autonomous driving scenarios.

\section{Dataset distribution}

To optimize the selection of representative and diverse data samples for memory replay, we employ a combination of TF-IDF feature extraction and K-means clustering. The process begins with vectorizing the textual content (e.g., questions) from the dataset using TF-IDF, which extracts key semantic patterns in the data. Following this, K-means clustering is applied to categorize the textual data into distinct clusters, ensuring diverse representation across varying data contexts. This pipeline helps maintain semantic richness and diversity in the selected memory samples.

To further refine the memory replay strategy, we dynamically calculate the number of samples to be selected from each cluster based on the proportional size of the cluster in the dataset. This ensures a balanced contribution of each cluster to the memory while reflecting their significance in the data. The sample calculation for each cluster is determined using the formula:
\begin{equation}
S_{c,n} = \left\lfloor \frac{|D_{c,n}|}{\sum_{i=1}^{k_n} |D_{i,n}|} \times S_n \right\rfloor,
\end{equation}
where \( D_{c,n} \) represents the volume of data for category \( c \) in task \( n \), \( k_n \) represents the total number of categories in that task, and \( S_n \) represents the total number of samples chosen. This approach ensures that the sampling process is proportional, improving the adaptability and robustness of the memory replay.

\subsection{Perception}
For the perception task, the TF-IDF and K-means pipeline generated five clusters with distinguishable thematic distributions. Each cluster encapsulates a specific aspect of the perception data. Key semantic themes were extracted for each cluster to provide a clear understanding of their focus areas. Some clusters focuses on observed objects and their statuses (e.g., "status object" "observed status"), some clusters predominantly describes side objects and their interactions and some highlight scene-level reasoning and positioning(e.g., "current scene" "relative positioning").

\begin{figure}
    \centering
    \includegraphics[width=\linewidth]{ 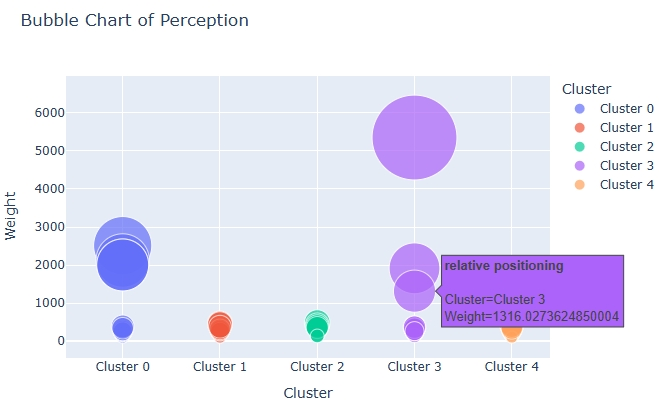}
    \caption{Bubble chart visualization of the perception task clusters generated using the TF-IDF and K-means pipeline.}
    \label{Perception_bubble}
\end{figure}

The bubble chart visualizes the clusters and their respective phrase frequencies. Each bubble represents a cluster, with its size proportional to the TF-IDF weight of the most representative phrases. In figure~\ref{Perception_bubble}, Cluster 3 ("relative positioning" 1316.03) captures crucial reasoning about positioning, which is essential for understanding the spatial relationships between objects in the driving environment. The visualization demonstrates how the clustering process preserves diversity and ensures that key semantic patterns are retained.

\subsection{Prediction}
For the Prediction task, the TF-IDF and K-means pipeline generated five clusters, each with distinct thematic distributions. These clusters capture different semantic aspects of the Prediction task.Some clusters focus on the ego vehicle state and related actions, some Highlight objects that could influence future states (e.g., "future state") and some emphasize decision-making semantics (e.g., "relevant decision").

\begin{figure}
    \centering
    \includegraphics[width=\linewidth]{ 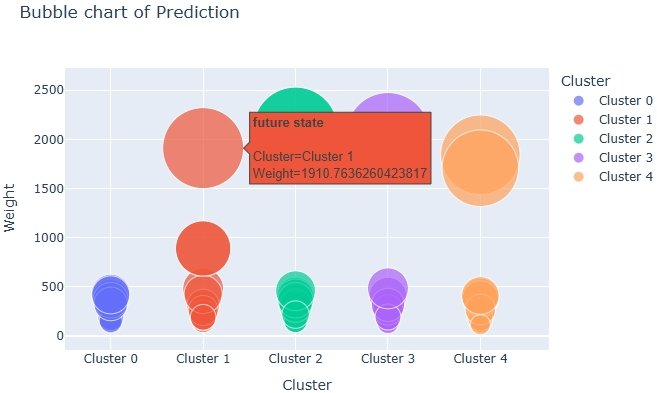}
    \caption{Bubble chart visualization of the prediction task clusters generated using the TF-IDF and K-means pipeline.}
    \label{Prediction_bubble}
\end{figure}

In figure~\ref{Prediction_bubble}, the phrase “future state” (1910.76) in Cluster 1 signifies the core semantic importance of predicting future scenarios. 

\subsection{Planning}
For the Planning task, the TF-IDF and K-means pipeline generated five clusters, each representing distinct semantic aspects of planning-related data.Some clusters highlight safe and dangerous actions in various scenarios (e.g., "safe actions" "dangerous actions"), some focus on vehicle considerations and actions taken to avoid collisions (e.g., "vehicle consider" "lead collision").

\begin{figure}
    \centering
    \includegraphics[width=\linewidth]{ 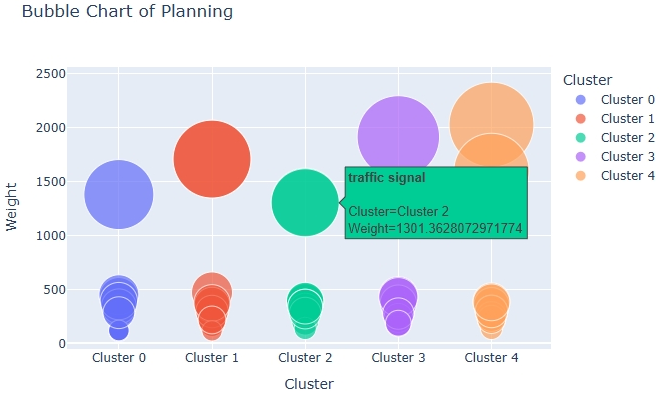}
    \caption{Bubble chart visualization of the planning task clusters generated using the TF-IDF and K-means pipeline.}
    \label{Planning_bubble}
\end{figure}

In figure~\ref{Planning_bubble}, the phrase “traffic signal” (1301.36) in Cluster 2 highlights the importance of considering external traffic signals in the planning process. 

\section{Case Analysis}
\label{sec: Case Analysis}
We illustrate our approach to multitask learning within an autonomous driving framework through multiple cases, categorizing results into correct and failure cases, with separate illustrations provided for each. The model is evaluated on previously trained tasks to assess the efficacy of continual learning methodologies.

\subsection{Correct Cases} 
The "Correct Cases" illustrate the effectiveness of our continual learning approach in retaining task-specific knowledge across multiple tasks. Each case includes corresponding images, a question related to task, the ground truth (GT) answer, and test results obtained after training on various tasks. The results are compared between models with and without continual learning to highlight the differences.

\begin{figure*}
    \centering
    \includegraphics[width=\linewidth, ]{ 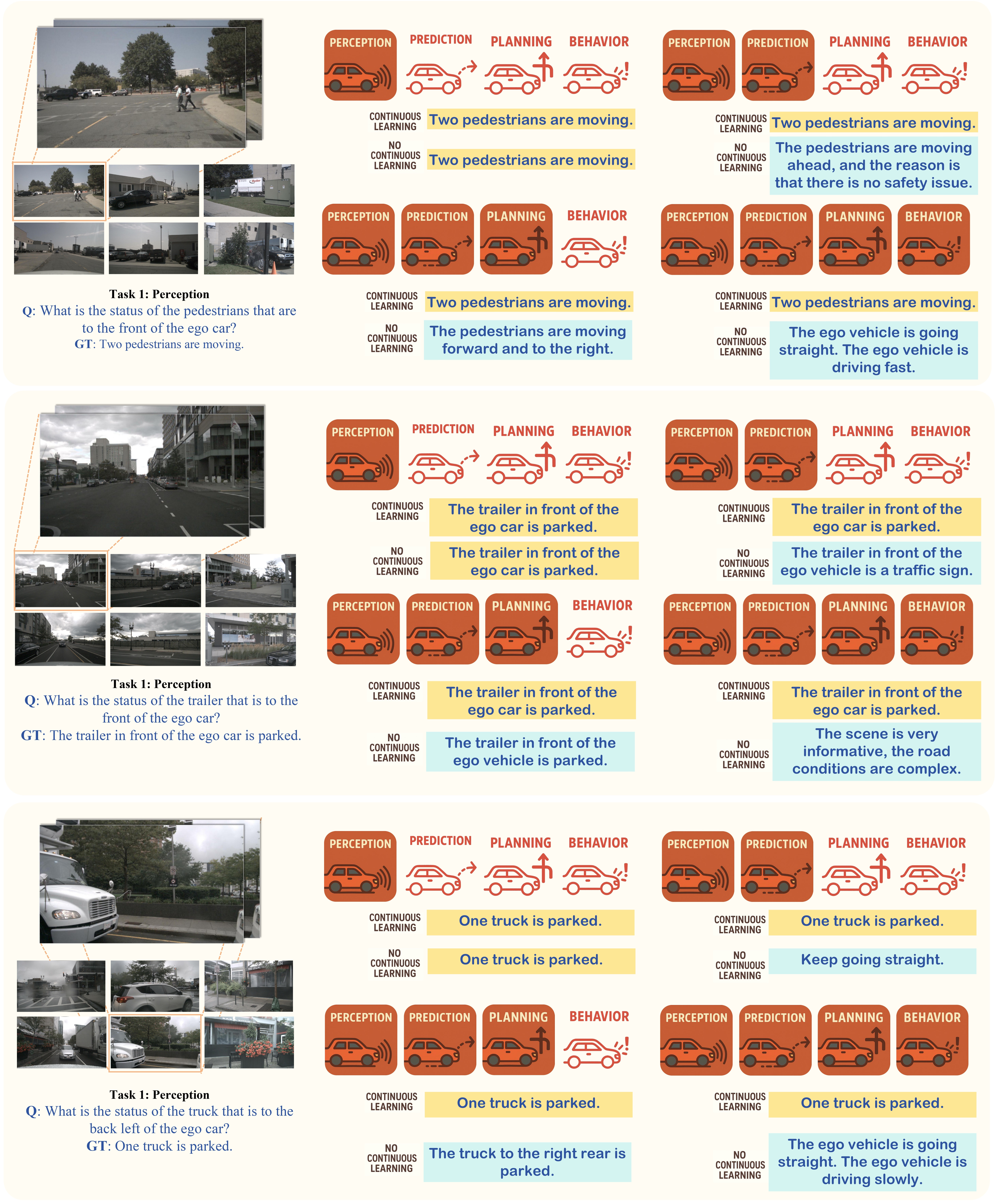}
    \caption{Correct Cases 1-3}
    \label{combine_1}
\end{figure*}

For example, in case 1 (Figure~\ref{combine_1}) the perception task focuses on identifying the status of pedestrians in front of the ego vehicle. The ground truth specifies that "two pedestrians are moving." The model equipped with continual learning methods successfully retains the ability to provide correct answers across all subsequent task tests, including perception, prediction, planning, and behavior tasks. For example, the output consistently states, "Two pedestrians are moving," regardless of the task being tested. This demonstrates the model's ability to preserve its perception capabilities even after multitask training.
In contrast, the model without continual learning exhibits catastrophic forgetting. While it correctly answers the perception task in isolation, multitask training leads to the gradual degradation of its perception capabilities. For instance, the outputs become task-dependent and deviate from the original ground truth. The prediction task output adds irrelevant reasoning about safety, the planning task shifts focus to spatial orientation, and the behavior task entirely disregards pedestrian status, instead describing the vehicle's movement. This underscores the challenge of knowledge retention in multitask learning without continual learning strategies.

\begin{figure*}
    \centering
    \includegraphics[width=\linewidth, ]{ 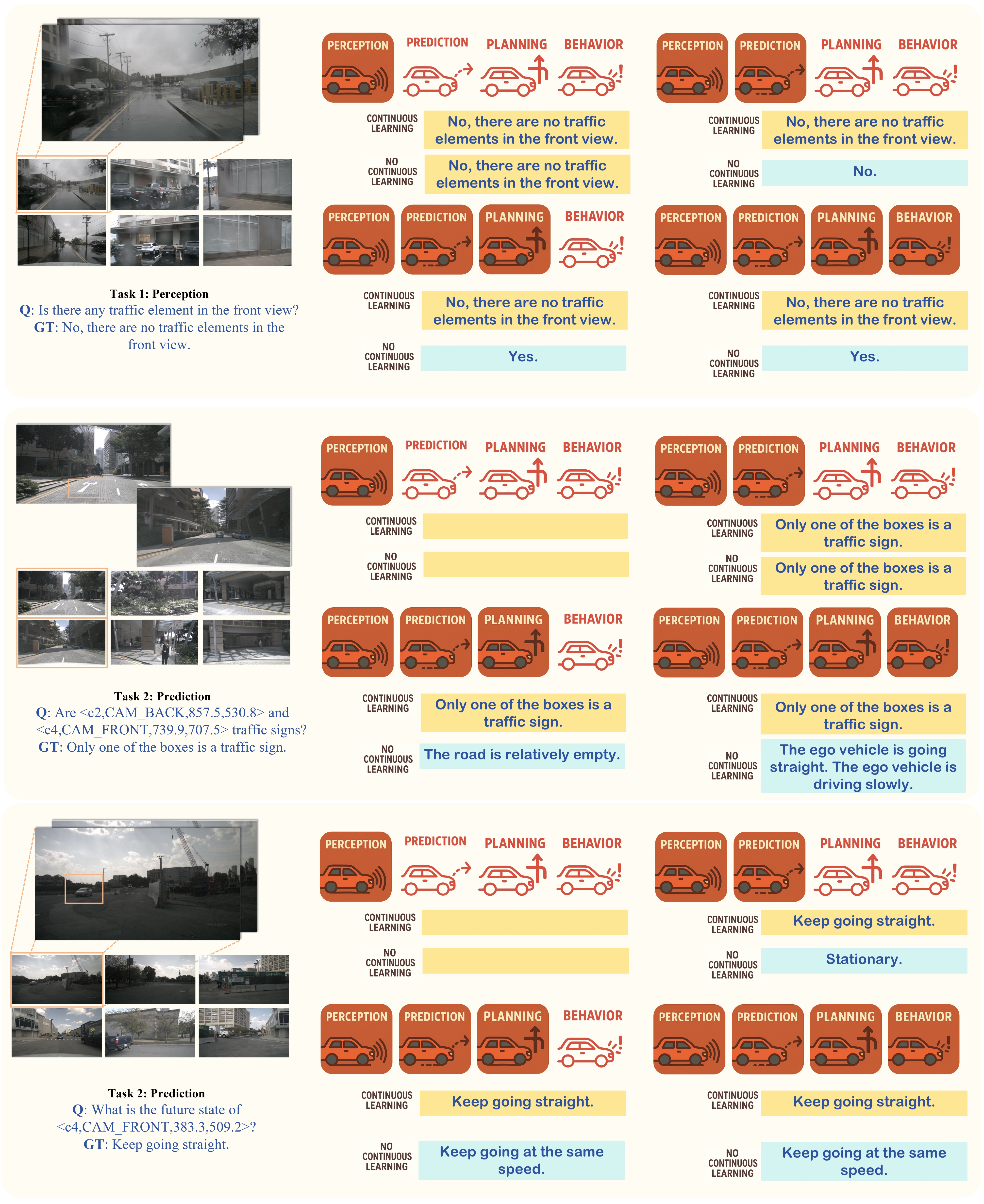}
    \caption{Correct Cases 4-6}    
    \label{combine_2}

\end{figure*}

In correct case 5 (Figure~\ref{combine_2}) examines the model's ability to determine whether two specified objects in the back and front cameras are traffic signs. The ground truth states that ``only one of the boxes is a traffic sign.'' Results, including images, the question, and the ground truth (GT), are compared for models with and without continual learning.
The model with continual learning retains task-specific knowledge effectively. Across all tasks, including prediction, planning, and behavior, it consistently outputs the correct answer: ``Only one of the boxes is a traffic sign'' This demonstrates the success of continual learning in preserving knowledge across tasks, even as the model undergoes additional training.
In contrast, the model without continual learning suffers from catastrophic forgetting. While it correctly identifies the traffic sign in the prediction task, its outputs for subsequent tasks deviate significantly, focusing on irrelevant observations (e.g., "The road is relatively empty") or providing unrelated descriptions (e.g., "The ego vehicle is driving slowly"). These inconsistencies underscore the necessity of continual learning to maintain knowledge across tasks and prevent forgetting.


\subsection{Failure Cases}
The "Failure Cases" highlight scenarios where the model's outputs deviate from the ground truth despite undergoing training with continual learning methods. Each case includes corresponding images, a question related to the task, the ground truth (GT), and test results after training on multiple tasks. 

\begin{figure*}
    \centering
    \includegraphics[width=\linewidth,]{ 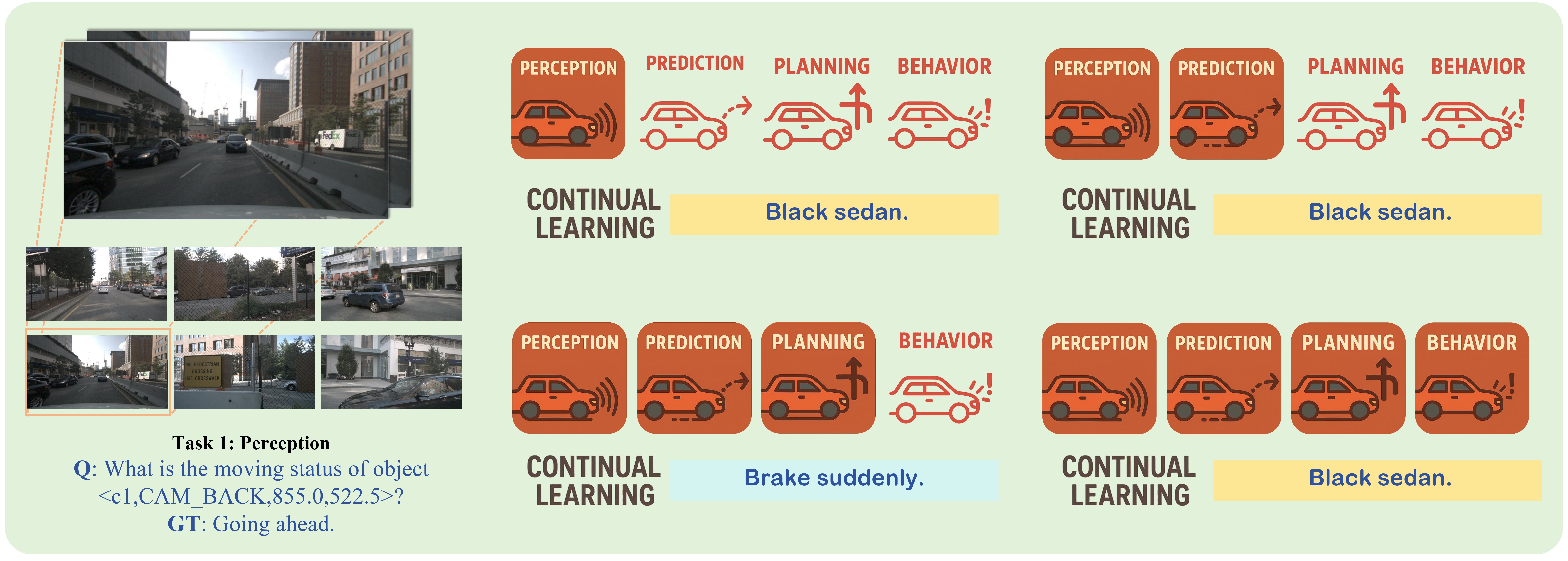}
    \caption{Failure Case 1}
    \label{fcase_1}
\end{figure*}

In failure case 1 (Figure~\ref{fcase_1}), the model is tasked with identifying the moving status of an object observed through the rear camera. The ground truth specifies "Going ahead." While the model correctly identifies the object's movement in the perception task, the outputs for subsequent tasks deviate. For example, the planning task incorrectly identifies the object's status as "Brake suddenly," likely due to the ambiguity in the dataset where the object's movement may appear inconsistent across frames. Such variations can introduce noise, leading to misinterpretation in planning tasks.

\begin{figure*}
    \centering
    \includegraphics[width=\linewidth]{ 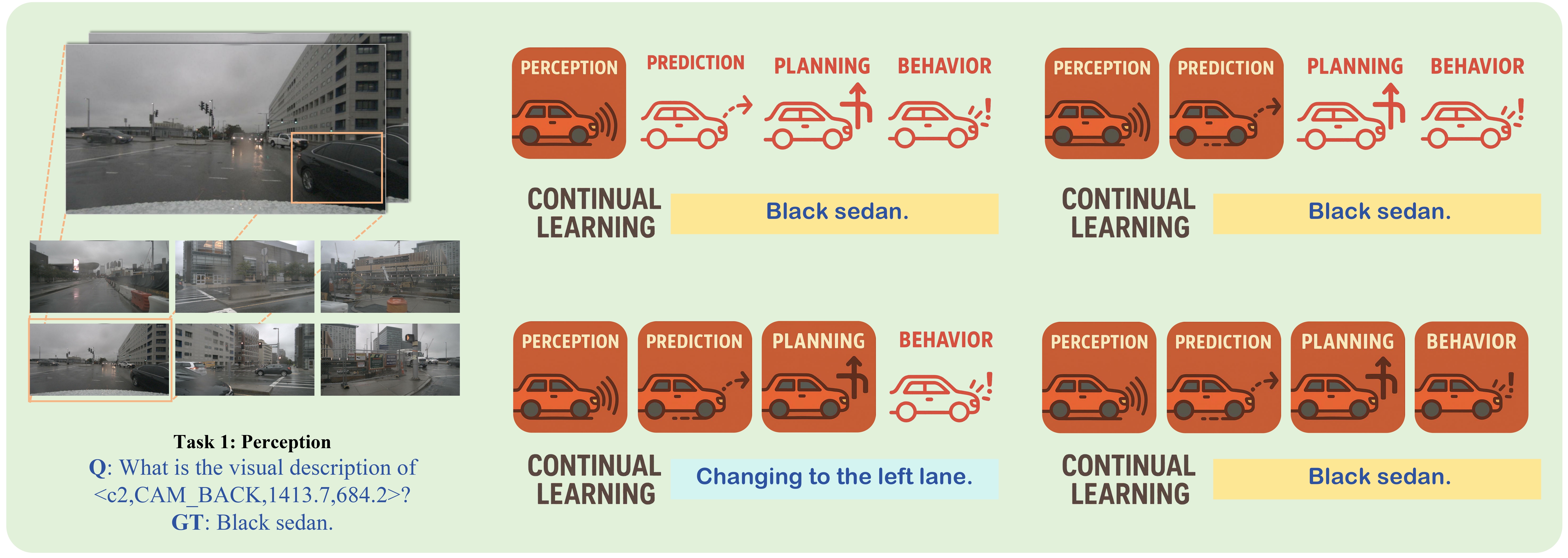}
    \caption{Failure Case 2}
    \label{fcase_2}
\end{figure*}

Failure case 2 (Figure~\ref{fcase_2}) evaluates the model’s ability to describe the visual characteristics of an object observed through the rear camera. The ground truth specifies "Black sedan." While the model correctly identifies the object as a "Black sedan" in the perception task, its outputs in subsequent tasks deviate. For example, the planning task concludes "Changing to the left lane," and the behavior task outputs "No entry." These inconsistencies could stem from the complexity of the scene, such as the presence of multiple objects or occlusions affecting the model's global understanding of the environment. Additionally, adverse weather conditions (e.g., low light or rain) in the images might increase the difficulty of the task and interfere with subsequent outputs.

\begin{figure*}
    \centering
    \includegraphics[width=\linewidth]{ 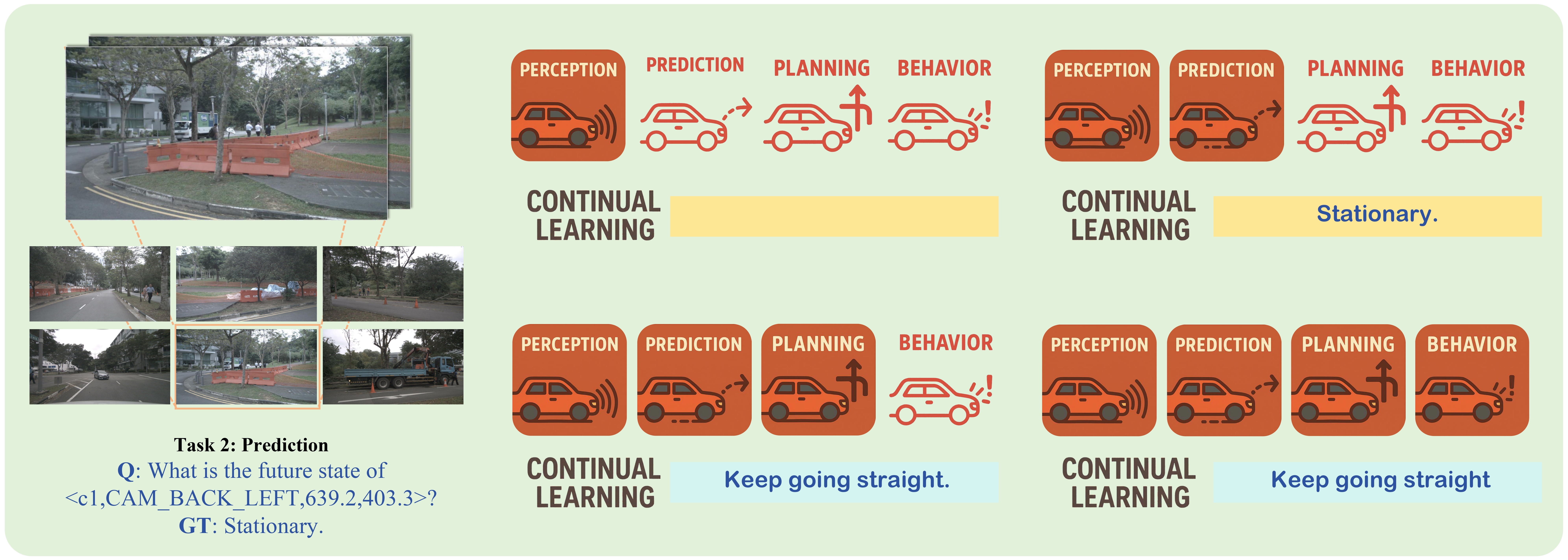}
    \caption{Failure Case 3}
    \label{fcase_3}
\end{figure*}

Failure case 3 (Figure~\ref{fcase_3})  focuses on the model's ability to predict the future state of an object located in the back-left camera view. The ground truth specifies "Stationary." While the model correctly identifies the object as "Stationary" in the prediction task, its outputs in subsequent tasks describe the object as "Keep going straight." This inconsistency could be influenced by the background context of the scene, where the object might appear near a dynamic environment (e.g., close to a main road), making it more likely to be perceived as moving. Additionally, the stationary state of the object might not be visually obvious in the images, particularly if it is positioned at an intersection or surrounded by other moving objects, leading to misinterpretation of its future state.

\begin{figure*}
    \centering
    \includegraphics[width=\linewidth, ]{ 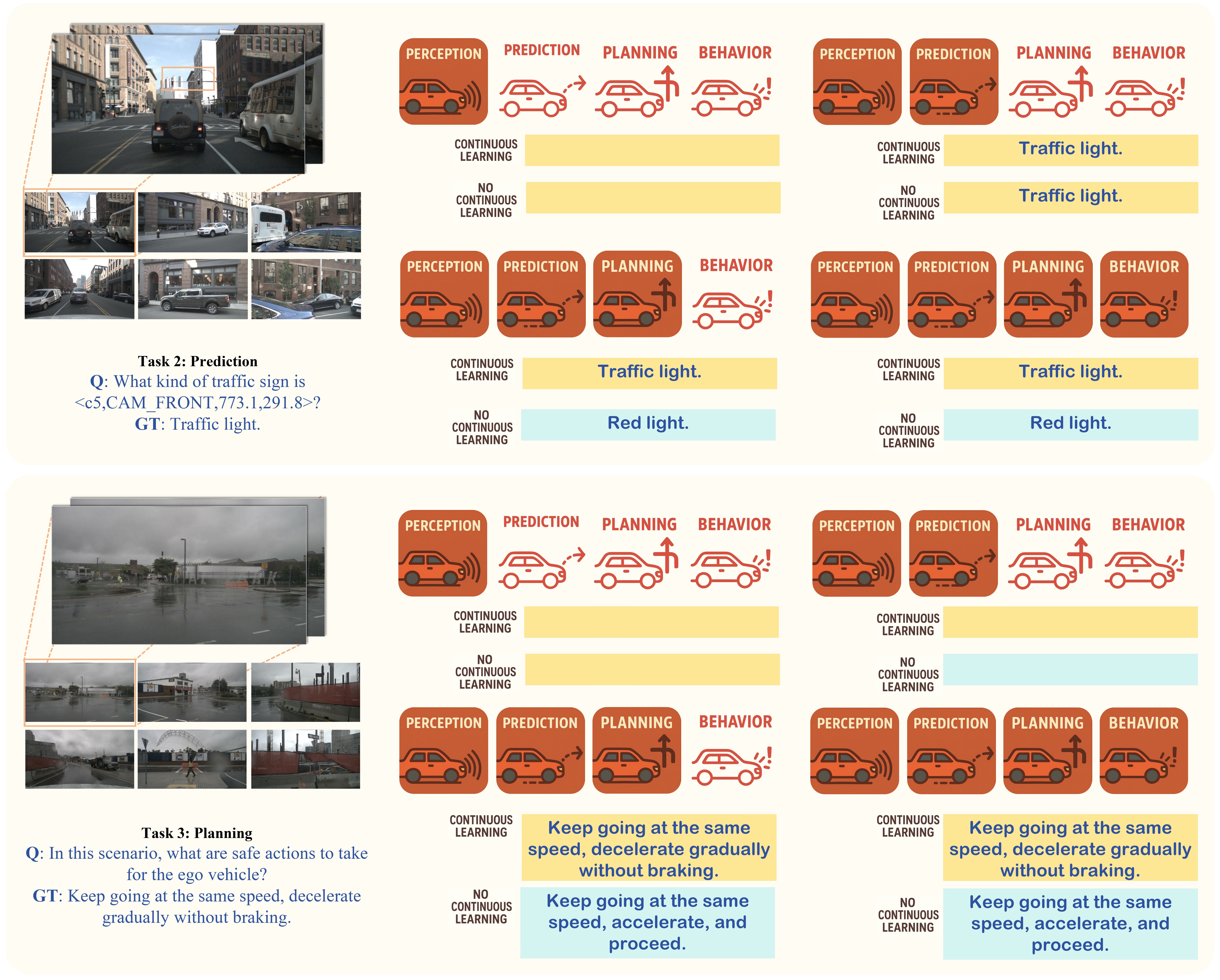}
    \caption{Correct Cases 7-8}
    \label{combine_3}
\end{figure*}

